\documentclass{article}

\usepackage{arxiv}

\usepackage[utf8]{inputenc} 
\usepackage[T1]{fontenc}    
\usepackage{hyperref}       
\usepackage{url}            
\usepackage{booktabs}       
\usepackage{amsfonts}       
\usepackage{nicefrac}       
\usepackage{microtype}      
\usepackage{lipsum}		
\usepackage{graphicx}
\usepackage{natbib}
\usepackage{doi}

\usepackage[nocompress]{cite}
\newtheorem{definition}{Definition}

\usepackage{times}
\usepackage{booktabs} 
\usepackage{paralist}
\usepackage{multirow}
\newcommand{\cut}[1]{}
\usepackage{graphicx}
\usepackage{color}
\usepackage{amsmath}
\usepackage{amssymb}

\usepackage{url}
\usepackage{wrapfig}
\usepackage{algpseudocode}
\usepackage{algorithm}
\usepackage{algcompatible}
\usepackage{epstopdf}
\usepackage{multirow}
\usepackage{rotating}
\usepackage{relsize}
\usepackage{appendix}
\usepackage{bbm}
\usepackage{mathtools}

\title{Enhancing Actionable Formal Concept Identification with Base-Equivalent Conceptual-Relevance}

\author{ \href{https://orcid.org/0000-0000-0000-0000}{\includegraphics[scale=0.06]{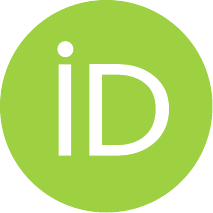}\hspace{1mm}Ayao Bobi}\\
	Department of Computer Science and Engineering \\ 
         University of Quebec in Outaouais \\
         Gatineau, Quebec, Canada\\
	\texttt{boba02@uqo.ca} \\
	\And
	\href{https://orcid.org/0000-0001-7410-4177}{\includegraphics[scale=0.06]{orcid.pdf}\hspace{1mm}Rokia Missaoui} \\
	Department of Computer Science and Engineering \\ 
         University of Quebec in Outaouais \\
         Gatineau, Quebec, Canada\\
	\texttt{rokia.missaoui@uqo.ca} \\
       \And
	\href{https://orcid.org/0000-0002-0604-2709}{\includegraphics[scale=0.06]{orcid.pdf}\hspace{1mm}Mohamed Hamza Ibrahim}\thanks{Corresponding author: mohamed.ibrahim@polymtl.ca}  \\
	Department of Computer Science and Engineering \\ 
         University of Quebec in Outaouais \\
         Gatineau, Quebec, Canada\\
	\texttt{mohamed.ibrahim@polymtl.ca} \\
}

\renewcommand{\shorttitle}{\textit{arXiv} Template}

\begin{document}
\maketitle

\begin{abstract}
In knowledge discovery applications, the pattern set generated from data can be tremendously large and hard to explore by analysts. In the Formal Concept Analysis (FCA) framework, there have been studies to identify important formal concepts through the stability index and other quality measures. In this paper, we introduce the \textit{Base-Equivalent Conceptual Relevance} ($\mathcal{BECR}$) score, a novel conceptual relevance interestingness measure for improving the identification of actionable concepts. From a conceptual perspective, the base and equivalent attributes are considered meaningful information and are highly essential to maintain the conceptual structure of concepts. Thus, the basic idea of $\mathcal{BECR}$ is that the more base and equivalent attributes and minimal generators a concept intent has, the more relevant it is. As such, $\mathcal{BECR}$ quantifies these attributes and minimal generators per concept intent. Our preliminary experiments on synthetic and real-world datasets show the efficiency of $\mathcal{BECR}$ compared to the well-known stability index.

\end{abstract}


\section{Introduction}
It is generally accepted, and with good reason, that we are ``drowning in data but starving for knowledge''. However, data mining and machine learning techniques may lead to a large set of patterns even when they are extracted from a small data collection. A few knowledge pieces are relevant to analysts and need to be identified among the generally large extracted pattern sets. Hence, the objective of this paper is to select ``conceptually relevant'' clusters using Formal Concept Analysis where any cluster is a formal concept, i.e., a node of a concept lattice.

Many concept selection strategies have been put forward in FCA literature, focusing on quantifying how important formal concepts are, with a detailed survey of interestingness measures for both concepts and association rules provided in \citep{KuznetsovInterest2018}. The stability index, first introduced in \citep{kuznetsov2007stability}, has long been considered one of the most viable interestingness measures. A further promising measure for evaluating concept quality was introduced later on, known as the Conceptual Relevance $\mathcal{CR}$ Index \citep{ibrahim2021conceptual}. While the first variant of $\mathcal{CR}$ takes advantage of the fact that using relevant attributes and concepts' minimal generators often leads to more interesting implications and association rules, there are still ways to make $\mathcal{CR}$ better for pattern mining by using other intrinsic properties of the formal context and the corresponding concept lattice.

In \citep{Ganter1999}, the notions of a base point and an extremal point of a given concept's intent or extent were described as a way to express its''indispensability''. The equivalent attributes within the concept, on the other hand, are frequently important local pieces of information and frequently lead to the extraction of meaningful implications and association rules from the concepts. Using this as inspiration, we present the Base-Equivalent $\mathcal{CR}$ ($\mathcal{BECR}$) index, a novel interestingness measure for improving the identification of actionable concepts. $\mathcal{BECR}$ quantifies three key elements in the concept intent: base attributes, equivalent attributes, and minimal generators, so the more of these elements there are in a concept, the more relevant it is.





The rest of the paper is organized as follows: Section~\ref{Back} recalls some basic definitions of FCA and stability index, while Section~\ref{BaseCR} describes our ($\mathcal{BECR}$) score to quantify a concept's relevance. In Section~\ref{Exp}, we conduct a preliminary experimental study with a discussion. Finally, Section~\ref{Conc} presents our conclusions.

\section{Background}
\label{Back} 
This section will briefly recall the basic notions of Formal Concept Analysis \citep{Ganter1999}] by using an illustrative example, which is a small formal context extracted from a product transaction database \citep{dong2005mining}. 

\begin{definition} [Formal context]
\label{cap:mydef:context-formel}
A formal context is a triple $\mathbb{K}= (\mathcal{G},\mathcal{M},\mathcal{I})$, where $\mathcal{G}$ is the set of objects, $\mathcal{M}$ is the set of attributes, and $\mathcal{I}$ is the set of binary relationships between $\mathcal{G}$ and $\mathcal{M}$ with $\mathcal{I} \subseteq \mathcal{G} \times \mathcal{M}$. The binary relationship $(g,m) \in I$ or $gIm$ holds if the object $g$ has attribute $m$.
\end{definition}

As shown in Table~\ref{cap:context-simple}, the illustrative example has a set of objects $\mathcal{G} = \{1,2,3,4,5\}$ that refer to transaction identifiers, and a set of attributes $\mathcal{M} = \{a,b,c,d,e,g,h,i\}$ that represent the products in the transactions.

\begin{definition} [Formal concept] The pair $c= (A,B)$ is a formal concept of $\mathbb{K}$ with extent $A \subseteq \mathcal{G}$ and intent $B \subseteq \mathcal{M}$ iff $B'=A$ and $A'=B$. Given the Galois connection in the lattice, $A'$ and $B'$ can be defined by the following derivation operators: 1) $A':=\{m \in M \mid g I m\ \forall g \in A\}$ which represent the attributes shared by objects in $A$. 2) $\ B':=\{g \in G \mid g I m\ \forall~m \in B\}$, which denotes the set of objects having all their attributes in $B$.

\end{definition}
For instance, the pair $c=(\{1,3,5\}, \{b,c,g,h,i\})$ is a formal concept, where $(\{1,3,5\})$ is its extent and $(\{b,c,g,h,i\})$ is its intent. We use $\mathcal{C}$ to denote all formal concepts extracted from $\mathbb{K}$. 

\begin{table}[!htbp]
\centering
\tabcolsep 3.5pt
\begin{tabular}{|l|l|l|l|l|l|l|l|l|l|l|l|l|l|l|}
\hline
\textbf{Id \& Product}  & \textbf{a} & \textbf{b} & \textbf{c} & \textbf{d} & \textbf{e} & \textbf{g} & \textbf{h} & \textbf{i} \\
\hline
\textbf{1} & X & X & X & X & X & X & X  & X  \\
\hline
\textbf{2} & X &   & X & X &   & X &    &   \\
\hline
\textbf{3} &   & X & X & X &   & X & X & X  \\
\hline
\textbf{4} & X & X &   & X &   &   & X & X \\
\hline
\textbf{5} &   & X & X &   & X & X & X & X \\
\hline
\end{tabular}\label{toycontext}
\caption{Formal context of the example \citep{dong2005mining}}
\label{cap:context-simple}
\end{table}

\begin{figure}[h]
\begin{center}\includegraphics[scale=0.40]{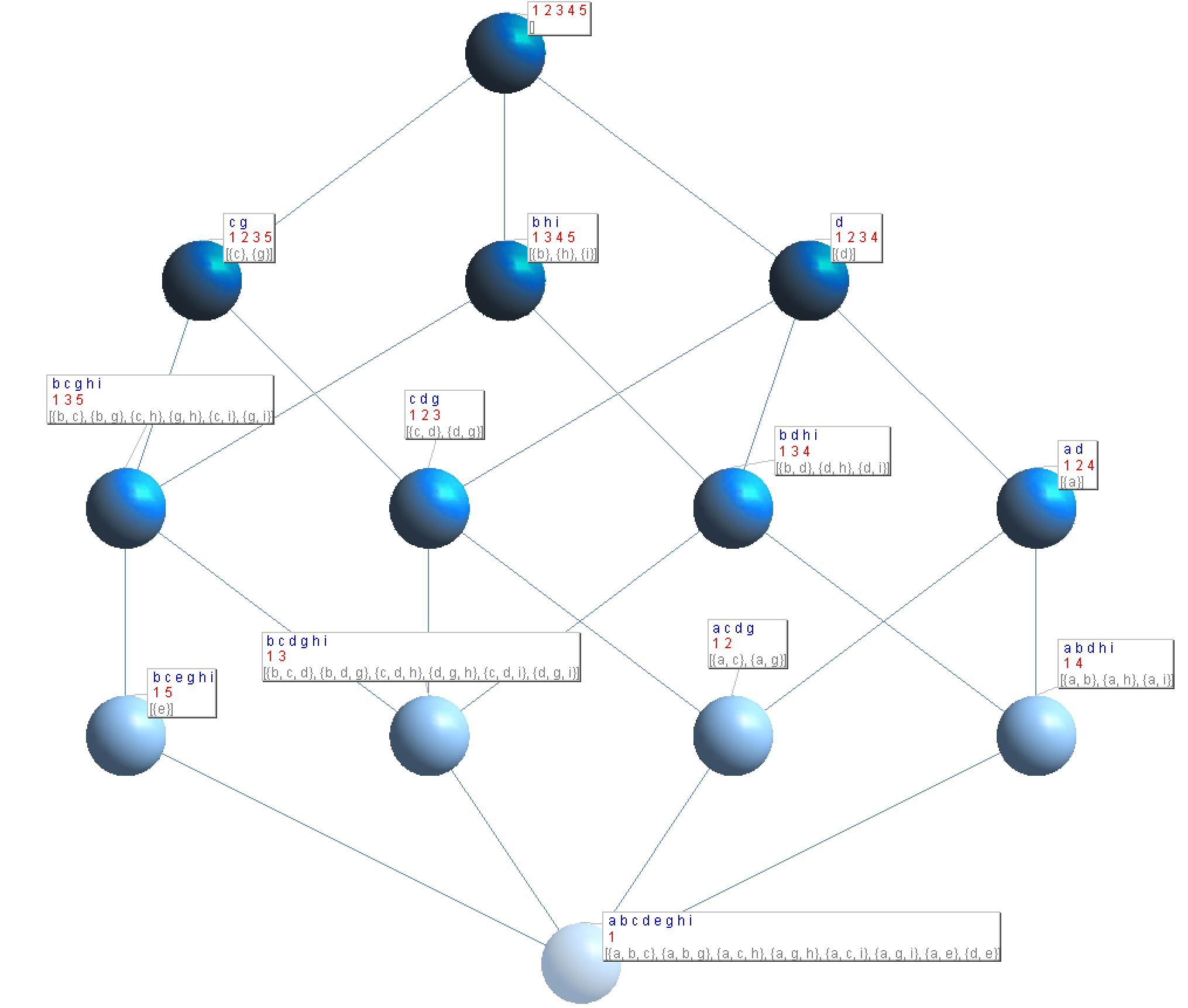}\end{center}
\caption{The concept lattice of the context in Table~\ref{cap:context-simple}.}
\label{exemple-treillis}
\end{figure}

The partial order $\preceq$ exists between two concepts: $c_i=(A_i,B_i) \preceq c_j=(A_j,B_j) \iff  A_i \subseteq A_j$ (and equivalently  $B_i \supseteq B_j$). The set $\mathcal{C}$ of all concepts together with their partial order form a concept lattice. Figure~\ref{exemple-treillis} shows the concept lattice corresponding to the illustrative context in Table~\ref{cap:context-simple}.



\begin{definition}[Minimal generator] \citep{Pfaltz2002}
\label{ming} 
A subset $h \subseteq B$ is a \textit{generator} of $B$ in a concept $c=(A,B)$ extracted from $\mathbb{K}$ iff $h^{\prime{}\prime{}}=B$. It is considered a \textit{minimal generator} when there does not exist any subset $h_j \subseteq h$ such that $h_j^{\prime{}\prime{}} =B$. We use $\mathcal{H}$ to denote the set of minimal generators of a concept $c$.
\end{definition}
For instance, the formal concept $(\{1, 2,3\},\{c, d, g\})$ has two minimal generators: $\mathcal{H}=\{\{c, d\}, \{d, g\}\}$ as shown in grey in Figure \ref{exemple-treillis}.

As defined below, the stability index $\sigma$ \citep{kuznetsov2007stability},is one prominent interestingness measure for assessing the actionability of concepts.

\begin{definition}[Stability Index] \citep{KuznetsovInterest2018}
Let $\mathbb{K} = (\mathcal{G},\mathcal{M},\mathcal{I})$ be a formal context and $c=(A,B)$ a formal concept of $\mathbb{K}$. The \textit{extensional stability} is: 
\begin{equation}\label{stability1}
    \sigma(c) = \frac{\mid \{e \in \mathcal{P}(B) | e^{\prime{}}= A\}\mid}{2^{|B|}}
\end{equation}
\end{definition}

where $\mathcal{P}(B)$ is the power set of $B$.  In Equation~\eqref{stability1}, the stability measures the strength of dependency between the extent $A$ and the attributes of the intent $B$. This measure quantifies the amount of noise that causes overfitting in the intent $B$. For example, the stability of the concept $(\{1, 2, 3, \},\{c, d, g\})$ is $\sigma=\frac{2}{2^3}=0.25$.

\section{Base-Equivalent Conceptual Relevance for Assessing Concept Quality}\label{BaseCR}

One goal toward accurately quantifying concept quality is to consider its relevant attributes and objects. As a result, the first step to enhancing $\mathcal{CR}$ index \citep{ibrahim2021conceptual} is to explore new types of conceptually relevant objects and attributes using base points \citep{Ganter1999} and equivalence property.


\begin{definition}[Equivalent attributes]
\label{equatt}
Given the formal concept $c=(A,B)$, the attributes $m_i,m_j \in B$, with $i \neq j$, are said to be equivalent w.r.t. $B$ if:
\begin{equation}
     m_i^{\prime{}} = m_j^{\prime{}} = A \text{ and } |m_i^{\prime{}\prime{}}|, |m_j^{\prime{}\prime{}}| > 1
\end{equation}
\end{definition}

That is, $m_i$ and $m_j$ are equivalent if their attribute concepts are identical. i.e., $c=\mu(m_i)=\mu(m_j)=(m_{i}', m_{i}'')$. This also means that these two equivalent attributes appear for the first time in the concept $c$ at earlier levels, beginning with the lattice's supremum, and the presence of one implies the presence of the other. For example, both attributes $\{\{c\},\{g\}\}$ are equivalent in the concept $(\{1, 2, 3, 5\},\{c,g\})$. Equivalent attributes are considered important and meaningful pieces of information in concept intent because it is widely known that they frequently result in the extraction of relevant association rules and implications from concepts.

\begin{definition}[Base attribute]
Given the formal concept $c=(A,B)$, the attribute $m \in B$ is said to be a base attribute of the intent $B$ if:
\begin{equation}\label{basepoint}
    m \notin \big( B \backslash \{ y \mid m^{\prime{}} \supseteq y^{\prime{}}\} \big)^{\prime{}\prime{}} 
\end{equation}
\end{definition}
This notion of base attribute is dual to base point of a given extent as defined in \citep{Ganter1999}.
That is, attribute $m$ is considered a base point of its concept intent $B$ if its presence is necessary to maintain the stability of the local conceptual structure of concept $c$. Taking it out of the concept's intent results in the loss of essential information or the collapse of the conceptual structure due to the expansion of its extent. From a conceptual standpoint, this means that the attribute $m$ represents relevant information and has a significant influence on the concept's intent. 

It is worth noting that the strict equality condition in Eq.~\eqref{basepoint} occurs when:
\begin{equation}
    m \notin \big( B \backslash \{ y \mid m^{\prime{}} = y^{\prime{}}\} \big)^{\prime{}\prime{}} 
\end{equation}
In this case, the base attribute turns into an extremal point, implying that it is highly essential to the conceptual intent. 

We have now paved the way for the Base-Equivalent Conceptual Relevance $\mathcal{BECR}$ index as follows:

\begin{equation}\label{crindex:01}
    \mathcal{BECR}(c) = \frac{1}{2} \big[\alpha(c) + \beta(c)\big]
\end{equation}
Where 
\begin{equation}
  \alpha(c) =
    \begin{cases}
      \frac{|\{m ~ \in ~ B \; \; \mid \; m \notin \big( B \backslash \{ y \mid m^{\prime{}} \supseteq y^{\prime{}}\} \big)^{\prime{}\prime{}}\}|}{|B|} & \text{if $B \not= \emptyset$}\\
      {\frac{|\{m~\in~B \; \mid \; |m^{\prime{}\prime{}}| > 1, m^{\prime{}} = A  \}|}{|B|}} & \text{if $B \not= \emptyset$ \text{ and } $m \in \big( B \backslash \{ y \mid m^{\prime{}} = y^{\prime{}}\} \big)^{\prime{}\prime{}}, \forall m \in B$}\\
      0 & \text{Otherwise.}
    \end{cases}   
    \label{alpha}
\end{equation}
and 
\begin{equation}\label{beta}
\beta(c) = \left\{
    \begin{array}{ll}
        Min(\frac{| \mathcal{H} |}{{|B|}}, 1) & \mbox{if } {|\mathcal{H}| > 1}\\
        \frac{1}{|B|} & \mbox{if }  {|\mathcal{H}| = 1}, {|h| < |B| } \mbox{ with } {h\in \mathcal{H}}\\
        0 & \mbox{otherwise.}
          
    \end{array}
\right.
\end{equation}

That is, in Eq.~\eqref{crindex:01}, the $\mathcal{BECR}(c)$ index computes the arithmetic average of $\alpha$ and $\beta$ relevance terms. In Eq.\eqref{alpha}, for the $\alpha$ term, we first iterate through the attributes of the intent B to count the number of the base attributes. Because the equivalent and base attributes are mutually exclusive, if there is no base attribute in the intent, we instead count the number of equivalent attributes by re-iterating through the attributes of the intent $B$. The size of the intent $|B|$ in the denominator serves as a normalization constant. As a result, this term quantifies the ratio of the conceptually relevant base and equivalent attributes in the concept intent B. In Eq.\eqref{beta}, for the $\beta$ term, we estimate the portions of minimal generators that result in relevant association rules. The efficient procedure called \textit{Minigen()}, proposed in \citep{ibrahim2021conceptual} can be used to calculate the minimal generator of the concept. We then count the number of minimal generators and divide by the size of the intent to scale the $\beta$ term value between $0$ and $1$. It is worth noting that the value of the $\alpha$ term can frequently be equal to zero if the intent does not contain neither  base nor equivalent attributes, while the $\beta$ term is often equal to $0$ when the intent is empty or the unique generator is equal to the concept intent. 

\begin{algorithm}[H]
\caption{Base-Equivalent Conceptual Relevance $\mathcal{BECR}$ index.}
\textbf{Input:}{ Concept $c=(A, B)$ and the set $\mathcal{H}$ of minimal generators of the intent $B$.} \\
\textbf{Output:}{ Base-Equivalent Conceptual Relevance $\mathcal{BECR}$ index.}
  \begin{algorithmic}[1] 
      \STATE $\alpha \leftarrow \beta \leftarrow 0$;  $\text{Count } \leftarrow 0$; 
      \STATEx \texttt{// Calculate $\alpha$ term.}
      \IF{$B \neq \emptyset$}
         
         \FOR{\text{each $m \in B$}}
            \IF{$m \notin \big( B \backslash \{ y \mid m^{\prime{}} \supseteq y^{\prime{}}\} \big)^{\prime{}\prime{}}$}
                \STATE $\text{Count } \leftarrow \text{Count}+1$
            \ENDIF
        \ENDFOR
		
		\ELSE 
         \FOR{\text{each $m \in B$}}
            \IF{$|m^{''}| > 1 \And m' = A$}
                \STATE $\text{Count } \leftarrow \text{Count}+1$;
            \ENDIF
        \ENDFOR
      \ENDIF
      \STATE $\alpha \leftarrow \frac{\text{Count}}{|B|}$
      
    \STATEx \texttt{// Calculate $\beta$ term.}
       \IF{$|\mathcal{H}|  > 1$}
         \STATE $ \beta \leftarrow Min( \frac{|\mathcal{H}|}{|B|}, 1)$;
         \ELSE 
         \IF{$|h| < |B|, h \in \mathcal{H}$}
         \STATE $ \beta \leftarrow\frac{1}{|B|}$;
        \ENDIF
        \ENDIF
       \STATE $\mathcal{BECR} \leftarrow (\alpha + \beta)/2$;
           \STATE $\text{\textbf{Return}} \; \mathcal{BECR}$;
    \end{algorithmic} 
    \label{CRalgo}
\end{algorithm}

The pseudo-code for the $\mathcal{BECR}$ index is provided by Algorithm~\ref{CRalgo}. The underlying concept and the set of minimal generators of its intent are fed as input into the algorithm. To demonstrate how it works, consider passing it the concept $c=(\{1,2,3\}, \{c,d,g\})$ with its minimal generator set $\{\{c,d\},\{d,g\}\}$. In lines 2–7, it begins by iterating through all attributes $\{c,d,g\}$ and counting the number of them that are base attributes. In this case, only the $\{d\}$ attribute satisfies the base attribute condition in line 4. Given that the base and equivalent attributes are mutually exclusive, the algorithm does not re-iterate through intent $B$ in lines 9–14 because it is supposed to look for equivalent attributes only if no base attributes are obtained. It then computes the value of $\alpha$, which is equal to $\frac{1}{3}$ for the provided concept in line 15. In lines 16–22, given that $\mathcal{H} = \{\{c,d\},\{d,g\}\}$, the algorithm proceeds with calculating the value of $\beta$, which is equal to $\frac{2}{3}$ for the illustrative concept. Line 25, it finally computes the $\mathcal{BECR}$ index by taking the arithmetic mean of the calculated $\alpha$ and $\beta$, which results in a $\frac{1}{2}$ value.



\section{Preliminary experiments}
\label{Exp}
We first selected the following commonly used synthetic and real-world datasets for evaluating relevancy scores: 1) \textit{Davis’s Shoutern Women} \citep{DavisGardner} describes a group of eighteen women from the American South who attended fourteen events; 2) \textit{CoinToss} \citep{felde2020null} is a random formal context generated by indirect Coin-Toss model generator; and 3) \textit{Phytotherapy} \citep{Phyto} which describes a set of diseases and medicinal plants that treat them. Table~\ref{table_stat} contains a description of the datasets. 

\begin{table}[!htbp]
\centering
\begin{tabular}{p{5cm}p{1.5cm}p{1.5cm}p{1.5cm}p{1.5cm}p{1.5cm}}
\noalign{\smallskip} \hline\noalign{\smallskip}
Dataset  & $|\mathcal{G}|$ & $|\mathcal{M}|$ &  $|\mathcal{I}|$  &  $|\mathcal{C}|$ & $\theta$ \\
\noalign{\smallskip} \hline\noalign{\smallskip}
\textit{Coin-Toss} & 793 & 10 & 3310  & 911 & 0.41\\
\noalign{\smallskip} \noalign{\smallskip}
\textit{Davis’s Shoutern Women} & 18   & 14 &  89 & 63 & 0.35\\
\noalign{\smallskip} \noalign{\smallskip}
\textit{Phytotherapy} & 142 & 108 &  511 & 302 & 0.03\\
\noalign{\smallskip} \hline\noalign{\smallskip}
\toprule\noalign{\smallskip}
\end{tabular}
\caption{A description of the tested datasets. $|\mathcal{G}|$ is the number of objects; $|\mathcal{M}|$ is the number of attributes; $|\mathcal{I}|$ is the number of links; $|\mathcal{C}|$ is the number of concepts; and $\theta$ is the context density.} 
\label{table_stat} 
\end{table}

We then conducted the following preliminary experimental evaluations to validate the performance of the $\mathcal{BECR}$ index. 
\subsection{Results and Discussions}
\paragraph{Experiment I.}
The purpose of this experiment was to compare $\mathcal{BECR}$ accuracy against the stability index empirically. That is, we attempted to compare the pairwise relevancy scores obtained by the $\mathcal{BECR}$ and stability indices for all concepts $\mathcal{C}$ extracted from each dataset. However, one common challenge was the lack of a clear assessment metric in the FCA literature that could be used directly for this purpose. Instead, we examined the linear correlation between the two indices empirically by calculating the Pearson correlation coefficient $\xi$ as:
\begin{equation}\label{pearson}
    \xi =  \frac{\sum_{i=1}^{|\mathcal{C}|} x_i y_i - |\mathcal{C}| \Bar{x}\Bar{y}}{\sqrt{(\sum_{i=1}^{|\mathcal{C}|} x_i^2 - |\mathcal{C}| \Bar{x}^2)} \sqrt{(\sum_{i=1}^{|\mathcal{C}|} y_i^2 - |\mathcal{C}| \Bar{y}^2)}}  
\end{equation}
Where 
$x_i$ and $y_i$ are the relevance scores obtained from $\mathcal{BECR}$ and stability indices for the concept $c_i \in \mathcal{C}$, respectively. $\bar{x} = \frac{1}{|\mathcal{C}|} \sum_{i=1}^{|\mathcal{C}|} x_i$ and $\Bar{y} = \frac {1} {|\mathcal{C}|}\sum_ {i = 1}^{|\mathcal{C}|} y_i$ are the average values of the relevance scores obtained from the two indices, respectively. It is worth noting that the linear relationship between two indices provides an indication of how the quantified relevance obtained by $\mathcal{BECR}$ is similar to or different from that obtained by stability.

\begin{figure}[!htbp]
  \begin{center}
          \includegraphics[width=70mm,height=45mm]{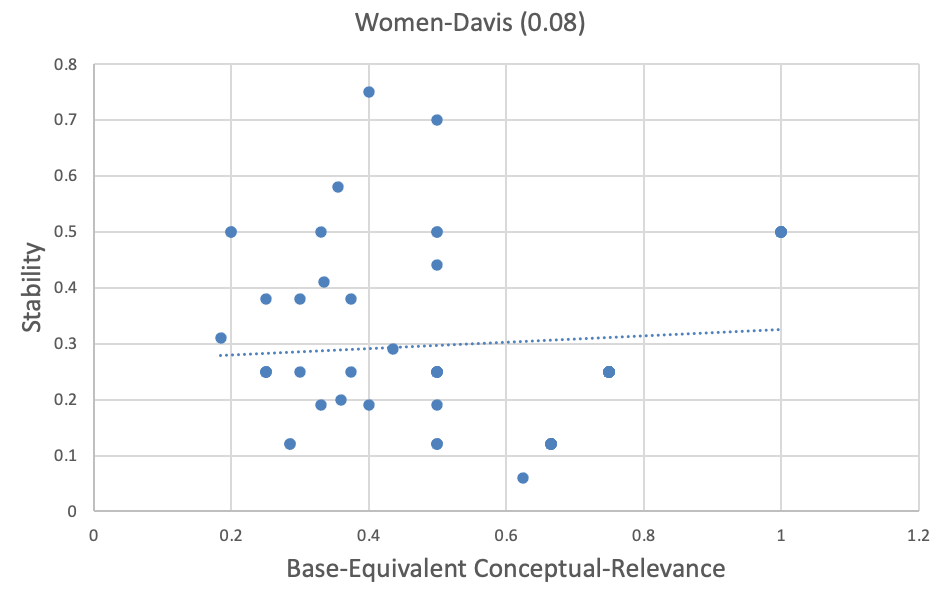} 
          \includegraphics[width=70mm,height=47mm]{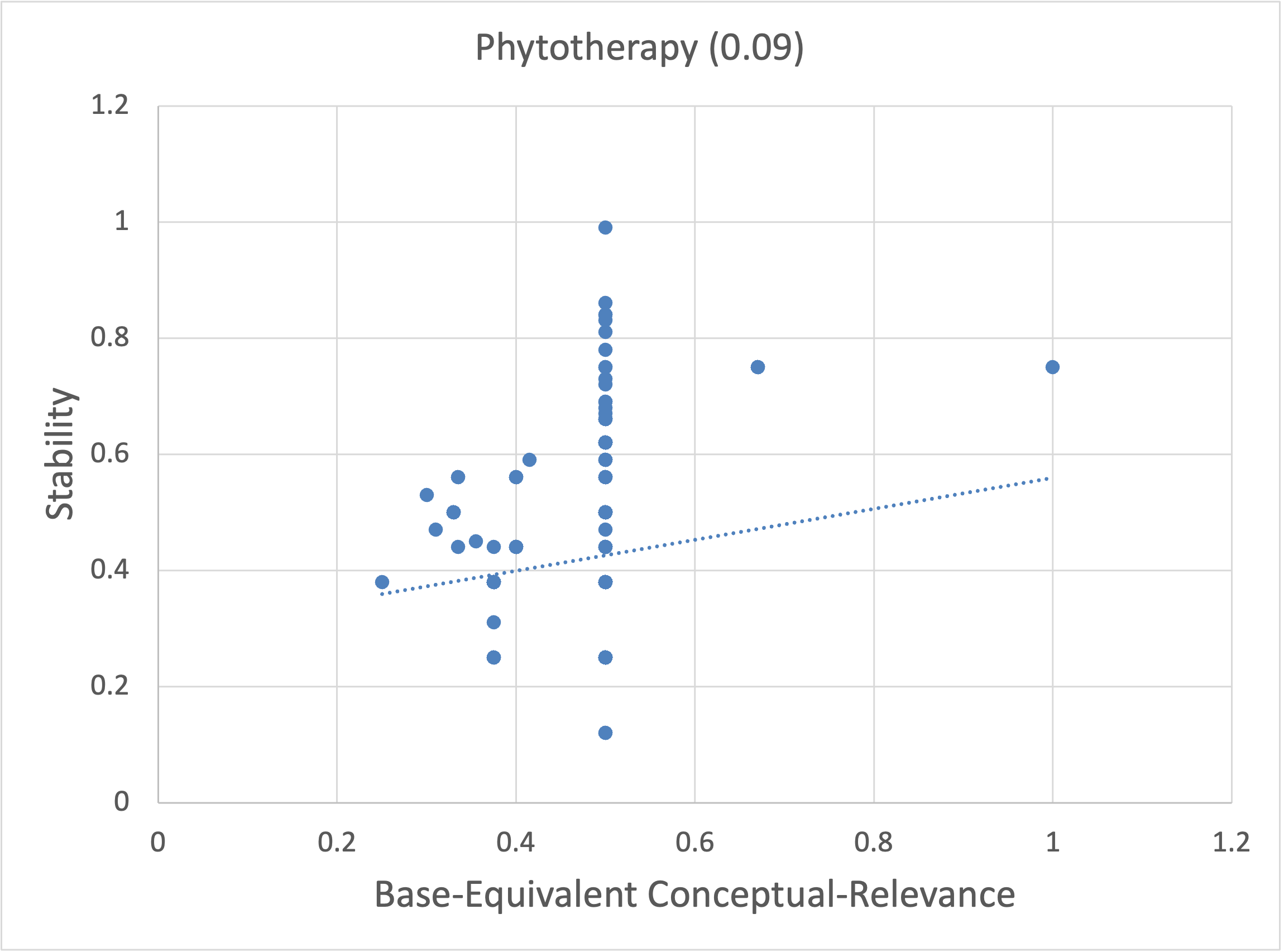}

          \includegraphics[width=70mm,height=47mm]{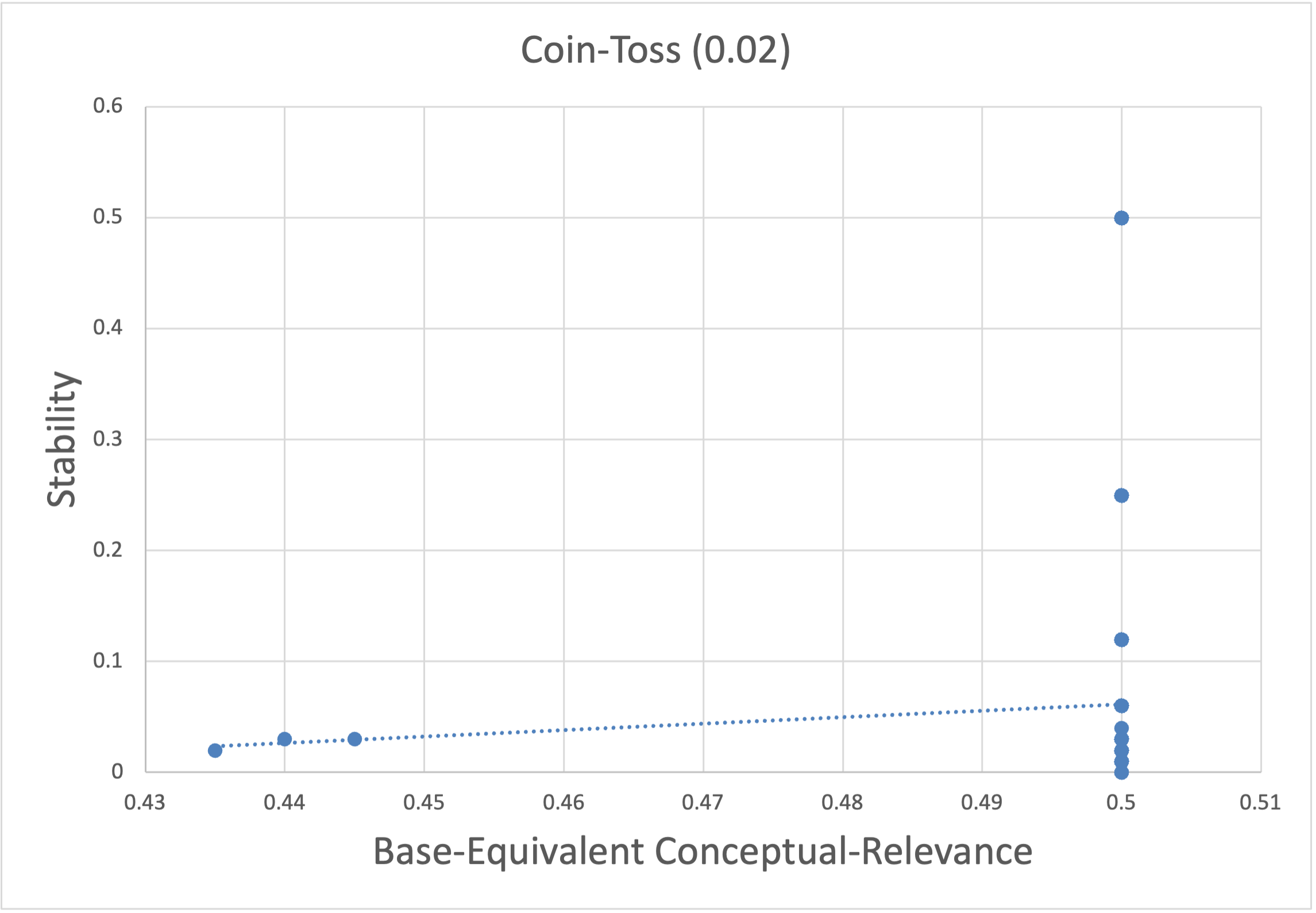}

    \end{center}
      \caption{The Pearson correlation scatter plot of $\mathcal{BECR}$ vs. stability on all concepts $\mathcal{C}$ extracted from the tested datasets. The Pearson correlation coefficient $\xi$ appears between parentheses.}
      \label{Exp1}
\end{figure}

Figure~\ref{Exp1} shows the Pearson correlation scatter plot of $\mathcal{BECR}$ and stability. Overall, the linear correlations between the two indices are very weak, with $0.08$, $0.09$ and $0.02$ Pearson correlation values on the Davis’s Shoutern Women, Phytotherapy, and Coin-Toss datasets, respectively. The experiment revealed the following findings: First, these weak linear correlations between the two indices indicate that the important quantities of $\mathcal{BECR}$—base and equivalent attributes, as well as minimal generators—capture certain relevant characteristics for assessing concept quality that stability does not. This ensures that the $\mathcal{BECR}$ index brings a new relevant aspect and has a unique trait for evaluating concept quality. Second, positive correlations are shown between both indices, indicating that when $\mathcal{BECR}$ produces high relevance scores for concepts, stability also tends to produce high relevance scores for similar concepts. Third, on Southern Woman-Davis, the loose clustering of the pairwise points around the best-fit line suggests that the two indices may have a nonlinear relationship. Fourth, stability produced many different relevancy scores for concepts, with relevance scores of $0.5$ on the phytotherapy and coin-toss datasets. This may indicate that stability might encounter difficulty accurately assessing the quality of concepts with relevance scores close to $0.5$, i.e., $0.5\pm \epsilon$, and the $\mathcal{BECR}$ index can provide better concept quality in this case.

\paragraph{Experiment II.}
The goal of this experiment was to investigate whether the $\mathcal{BECR}$ index is faster than stability. As a result, we compared the computational performance of both indices by computing their average execution time ($\tau$) as follows:
\begin{equation}\label{Timeeq}
    \tau =  \frac{\sum_{i=1}^n t_i}{n} 
\end{equation}
Where $t_{i}$ is the CPU time to calculate the relevance index of the concept $c_i$.

\begin{figure}[!htbp]
  \begin{center}
          \includegraphics[width=70mm,height=45mm]{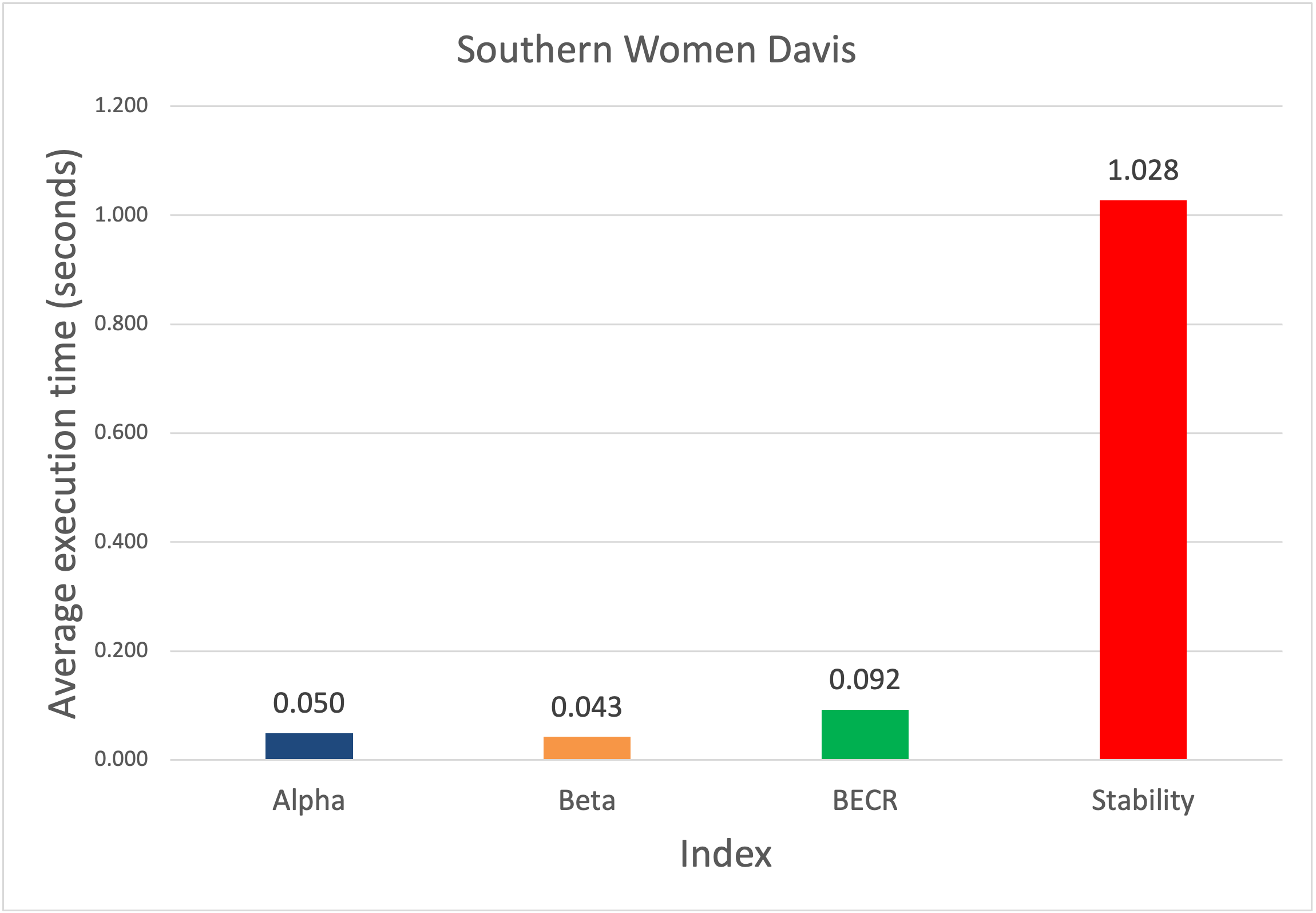} 
          \includegraphics[width=70mm,height=45mm]{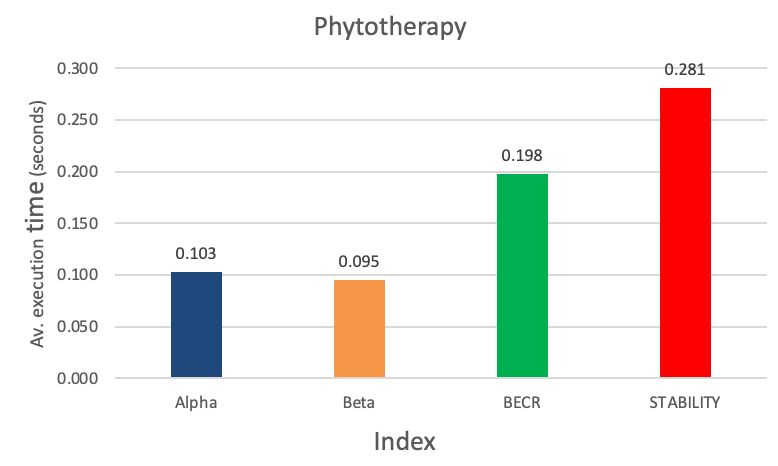}
          \includegraphics[width=70mm,height=45mm]{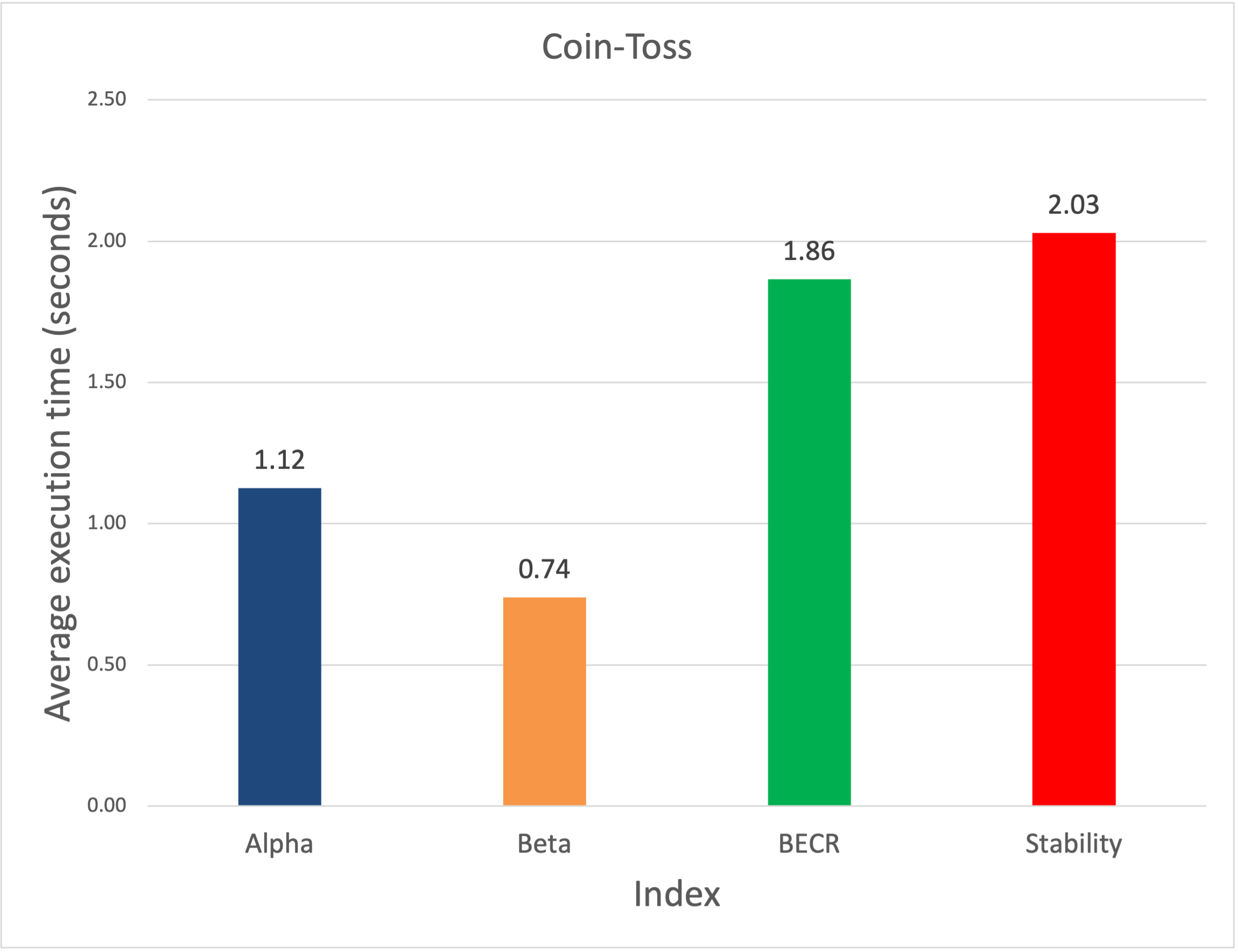}

    \end{center}
      \caption{The average execution time $\tau$ of $\mathcal{BECR}$ vs. stability on the tested datasets.}
      \label{Exp2}
\end{figure}

Figure~\ref{Exp2} shows the average execution time of $\mathcal{BECR}$ and stability indices. Overall, $\mathcal{BECR}$ outperformed stability across all datasets tested. It performed faster than the stability on the Phytotherapy and Coin-toss datasets by $29\%$ and $8\%$, respectively, and it dominated the stability on the Davis's Southern Women dataset by at least eleven times faster. From a computational perspective, this is due to the fact that the $\alpha$ term of $\mathcal{BECR}$ takes a polynomial amount of time, which depends on the number of base and equivelent attributes in the intent, and the $\beta$ term efficiently finds the set of minimal generators using the \textit{Minigen()} procedure \citep{ibrahim2021conceptual}, which also has a polynomial time. Stability, on the other hand, has an exponential time complexity because it might rely on verifying all subsets in the intent powerset or traversing deeply through the concept lattice to calculate generators.


\section{Conclusion}\label{Conc}
With the goal of advancing an efficient pattern mining process, the Base-Equivalent Conceptual-Relevance $\mathcal{BECR}$ index was introduced as a novel interestingness measure for identifying important formal concepts. The $\mathcal{BECR}$ index is distinctive in that it leverages base and equivalent attributes as new meaningful pieces of information in the concept that ought to be considered when assessing its relevance. Preliminary results on synthetic and real-world datasets showed that $\mathcal{BECR}$ was efficient compared to stability, with several notable findings: First, there are weak linear correlations between the $\mathcal{BECR}$ and stability indices. This indicates that the important quantities of $\mathcal{BECR}$—base, equivalent, and minimal generators—capture more relevant characteristics for concept quality assessment than the generators alone in the stability formula. Second, the positive correlations between both indices imply that when $\mathcal{BECR}$ produces high relevance scores, stability also yields high relevance scores for similar concepts. Finally, stability might have difficulty evaluating concepts with relevance scores close to $0.5$.



      

\bibliographystyle{unsrtnat}
\bibliography{ReferencesAyao}






\end{document}